\documentstyle[aclap]{article}

\title{Learning Methods for Combining Linguistic Indicators \\
to Classify Verbs}

\author{Eric V. Siegel \\
Department of Computer Science\\
Columbia University\\
New York, NY  10027\\
evs@cs.columbia.edu}

\begin{document}

\maketitle

%\noindent
%Content Areas: corpus-based methods, machine learning, natural language
%understanding 
%\smallskip

%\noindent
%Word Count: 3,484
%\smallskip

%\noindent
%Tracking Number: A333
%\smallskip

%\noindent
%This paper is not under review or accepted for publication in another
%conference or journal.

\begin{abstract}
%\begin{quote}

Fourteen linguistically-motivated numerical indicators are evaluated for
their ability to categorize verbs as either {\it states} or {\it events}.
The values for each indicator are computed automatically across a corpus of
text.  To improve classification performance, machine learning techniques
are employed to combine multiple indicators.  Three machine learning
methods are compared for this task: decision tree induction, a genetic
algorithm, and log-linear regression.

%\end{quote}
\end{abstract}

\section{Introduction}

The ability to distinguish states, e.g., {\it ``Mark seems happy,''} from
events, e.g., {\it ``Ren\'{e}e ran down the street,''} is a necessary
prerequisite for interpreting certain adverbial adjuncts, as well as
identifying temporal constraints between sentences in a discourse
\cite{moens,dorr,klavans2}.  Furthermore, {\it stativity} is the first of
three fundamental temporal distinctions that compose the {\it aspectual
class} of a clause.  Aspectual classification is a necessary component for
a system that analyzes temporal constraints, or performs lexical choice and
tense selection in machine translation
\cite{moens,passonneau,dorr,klavans2}.

Researchers have used empirical analysis of corpora to develop
linguistically-based numerical indicators that aid in aspectual
classification \cite{klavans,siegel4}.  Specifically, this technique takes
advantage of linguistic constraints that pertain to aspect, e.g., only
clauses that describe an event can appear in the progressive.  Therefore, a
verb that appears more frequently in the progressive is more likely to
describe an event.

In this paper, we evaluate fourteen quantitative linguistic indicators for
their ability to classify verbs according to stativity.  The values of
these indicators are computed automatically across a corpus of text.
Classification performance is then measured over an unrestricted set of
verbs.  Our analysis reveals a predictive value for several indicators that
have not traditionally been linked to stativity in the linguistics
literature.  Then, in order to improve classification performance, we apply
machine learning methods to combine multiple indicators.  Three machine
learning techniques are compared for this task: decision tree induction, a
genetic algorithm, and log-linear regression.

In the following sections, we further detail and motivate the distinction
between states and events.  Next, we describe our approach, detailing the
set of linguistic indicators, the corpus and tools used, and the machine
learning methods.  Finally, we present experimental results and discuss
conclusions and future work.

\section{Stative and Event Verbs}

Stativity must be identified to detect temporal constraints between clauses
attached with {\it when}.  For example, in interpreting, {\it ``She had
good strength {\bf when} objectively tested,''}\footnote{These examples of
{\it when} come from the corpus of medical discharge summaries used for
this work.} the {\it have}-{\bf state} began before or at the beginning of
the {\it test}-event, and ended after or at the end of the {\it
test}-event.  However, in interpreting, {\it ``Phototherapy was
discontinued {\bf when} the bilirubin came down to 13,''} the {\it
discontinue}-{\bf event} began at the end of the {\it come}-event.  As
another example, the simple present reading of an event, e.g., {\it ``He
jogs,''} denotes the {\it habitual} reading, i.e., {\it ``every day,''}
whereas the simple present reading of a state, e.g., {\it ``He appears
healthy,''} implies {\it ``at the moment.''}

Identifying stativity is the first step toward aspectually classifying a
clause.  Events are further distinguished by two additional features: 1)
telic events have an explicit culminating point in time, while non-telic
events do not, and 2) extended events have a time duration, while atomic
events do not.  Detecting the telicity and atomicity of a clause is
necessary to identify temporal constraints between clauses and to interpret
certain adverbial adjuncts \cite{moens,passonneau,dorr,klavans2}.  However,
since these features apply only to events and not to states, a clause first
must be classified according to stativity.

Certain features of a clause, such as adjuncts and tense, are constrained by
and contribute to the aspectual class of the clause
\cite{vendler,dowty,pustejovsky,passonneau,klavans2}.  Examples of such
constraints are listed in Table~\ref{klavans-table1}.  Each entry in this
table describes a syntactic aspectual {\it marker} and the constraints on
the aspectual class of any clause that appears with that marker.  For
example, a telic event can be modified by a duration {\it in}-PP, as in
{\it ``You found us there {\bf in ten minutes},''} but a state cannot,
e.g., {\it ``*You loved him {\bf in ten minutes}.''}

In general, the presence of these linguistic markers in a particular clause
indicates a constraint on the aspectual class of the clause, but the
absence thereof does not place any constraint.  This makes it difficult for
a system to aspectually classify a clause based on the presence or absence
of a marker.  Therefore, these linguistic constraints are best exploited by
a system that measures their frequencies across verbs.

Klavans and Chodorow \shortcite{klavans} pioneered the application of
%Klavans and Chodorow \cite{klavans} pioneered the application of
statistical corpus analysis to aspectual classification by placing verbs on
a ``stativity scale'' according to the frequency with which they occur in
the progressive.  This way, verbs are automatically ranked according to
their propensity towards stativity.  We have previously applied this
principle towards distinguishing telic events from non-telic events
\cite{siegel4}.  Classification performance was increased by combining
multiple aspectual markers with a genetic algorithm.

\begin{table}
\begin{center}
\begin{tabular}{||l|l||}	\hline
If a verb can occur:&...then it must be:\\
\hline
\hline
in the progressive&{\bf Extended Event}\\
\hline
with a temporal adverb &{\bf Event}\\
(e.g., {\it then})&\\
\hline
with a duration {\it in}-PP &{\bf Telic Event}\\
(e.g., {\it in an hour})&\\
\hline
in the perfect tense&{\bf Telic Event} or {\bf State}\\
\hline
\end{tabular}
\end{center}
\caption{Example linguistic constraints excerpted from
Klavans \protect{\shortcite{klavans2}}.} 
\label{klavans-table1}
\end{table}

\section{Approach}

Our goal is to exploit linguistic constraints such as those listed in
Table~\ref{klavans-table1} by counting their frequencies in a corpus.  For
example, it is likely that event verbs will occur more frequently in the
progressive than state verbs, since the progressive is constrained to occur
with event verbs.  Therefore, the frequency with which a verb occurs in the
progressive {\it indicates} whether it is an event or stative verb.

We have evaluated 14 such linguistic indicators over clauses selected
uniformly from a text corpus.  In this way, we are measuring classification
performance over an unrestricted set of verbs.  First, the ability for each
indicator to individually distinguish between stative and event verbs is
evaluated.  Then, in order in increase classification performance, machine
learning techniques are employed to combine multiple indicators.

In this section, we first describe the set of linguistic indicators used to
discriminate events and states.  Then, we show how machine learning is used
to combine multiple indicators to improve classification performance.
Three learning methods are compared for this task.  Finally, we describe
the corpus and evaluation set used for these experiments.

\subsection{Linguistic Indicators}

The first column of Table~\ref{marker-table} lists the 14 linguistic
indicators evaluated in this paper for classifying verbs.  The second and
third columns show the average value for each indicator over stative and
event verbs, respectively, as computed over a corpus of parsed clauses,
described below in Section~\ref{corpus}.  These values, as well as the third
column, are further detailed in Section~\ref{baba}.

Each verb has a unique value for each indicator.  The first indicator, {\tt
frequency}, is simply the the frequency with which each verb occurs.  As
shown in Table~\ref{marker-table}, stative verbs occur more frequently than
event verbs in our corpus.

The remaining 13 indicators measure how frequently each verb occurs in a
clause with the linguistic marker indicated.  This list includes the four
markers listed in Table~\ref{klavans-table1}, as well as 9 additional
markers that have not previously been linked to stativity.  For example,
the next three indicators listed in Table~\ref{marker-table} measure the
frequency with which verbs 1) are modified by {\it not} or {\it never}, 2)
are modified by a {\it temporal} adverb such as {\it then} or {\it
frequently}, and 3) have no deep subject (passivized phrases often have no
deep subject, e.g., {\it ``She was admitted to the hospital''}).  As
shown, stative verbs are modified by {\it not} or {\it never} more
frequently than event verbs, but event verbs are modified by {\it temporal}
adverbs more frequently than stative verbs.
For further detail regarding the set of 14
indicators, see Siegel \shortcite{siegel7}.

An individual indicator can be used to classify verbs by simply
establishing a threshold; if a verb's indicator value is below the
threshold, it is assigned one class, otherwise it is assigned the
alternative class.  For example, in Table~\ref{verb-table}, which shows the
predominant class and four indicator values corresponding to each of four
verbs, a threshold of 1.00\% would allow events to be distinguished from
states based on the values of the {\it not/never} indicator.  The next
subsection describes how all 14 indicators can be used together to classify
verbs.

\begin{table}
\begin{center}
\begin{tabular}{|l|r|r|r|}	\hline
Linguistic       & Stative & Event &T-test\\
Indicator & Mean &    Mean       &P-value\\
\hline
\hline
{\tt frequency} & {\tt 932.89} & {\tt 667.57} & 0.0000 \\ 
{\it ``not''} or {\it ``never''} & 4.44\% & 1.56\%& 0.0000 \\ 
{\it temporal} adverb & 1.00\% & 2.70\%& 0.0000 \\ 
no deep subject & 36.05\% & 57.56\%& 0.0000 \\ 
past/pres participle &  20.98\% & 15.37\%& 0.0005 \\ 
duration {\it in}-PP &  0.16\% & 0.60\% & 0.0018 \\ 
perfect &  2.27\% & 3.44\% &  0.0054 \\ 
\hline
present tense &  11.19\% & 8.94\%& 0.0901 \\ 
progressive &  1.79\% & 2.69\% & 0.0903 \\ 
{\it manner} adverb & 0.00\% & 0.03\%& 0.1681 \\ 
{\it evaluation} adverb &  0.69\% & 1.19\%& 0.1766 \\ 
past tense &  62.85\% & 65.69\%& 0.2314 \\ 
duration {\it for}-PP &  0.59\% & 0.61\%& 0.8402 \\ 
{\it continuous} adverb &  0.04\% & 0.03\%& 0.8438 \\ 
\hline
\end{tabular}
\end{center}
\caption{Indicators discriminate between two classes.}
\label{marker-table}
\end{table}

\begin{table*}
\begin{center}
\begin{tabular}{|l|r|r|r|r|r|} \hline
     &       &   & {\it ``not''} or & {\it temporal} & no deep   \\
Verb & class & {\tt freq} & {\it ``never''} & adverb          & subject \\
\hline
\hline
{\it show} &  state  & 2,131 & 1.55\% & 0.52\% & 18.07\% \\
{\it admit} & event  & 1,895 & 0.05\% & 1.11\% & 91.13\% \\ 
{\it discharge} & event & 1,608 & 0.50\% & 1.87\% & 96.64\% \\
{\it feel} & state & 1,177 & 4.61\%  &  1.20\% &  52.52\%   \\
\hline
\end{tabular}
\end{center}
\caption{Example verbs and their indicator values.}
\label{verb-table}
\end{table*}

\subsection{Combining Indicators with Learning}

Given a verb and its 14 indicator values, our goal is to use all 14 values
in combination to classify the verb as a state or an event.  Once a
function for combining indicator values has been established, previously
unobserved verbs can be automatically classified according to their
indicator values.  This section describes three machine learning methods
employed to this end.

\noindent
{\bf Log-linear regression.}  As suggested by Klavans and Chodorow
\shortcite{klavans}, a weighted sum of multiple indicators that results in
%\cite{klavans}, a weighted sum of multiple indicators that results in
one ``overall'' indicator may provide an increase in classification
performance.  This method embodies the intuition that each indicator
correlates with the probability that a verb describes an event or state,
but that each indicator has its own unique scale, and so must be weighted
accordingly.  One way to determine these weights
is log-linear regression
\cite{santner}, a popular technique for binary classification.  This
technique, which is more extensive than a simple weighted sum, applies an
inverse logit function, and employs the iterative reweighted least squares
algorithm \cite{baker}.

\noindent
{\bf Genetic programming.}  An alternative to avoid the limitations of a
linear combination is to generate a non-linear function tree that combines
multiple indicators.  A popular method for generating such function trees
is a genetic algorithm \cite{holland,goldberg}.  The use of genetic
algorithms to generate function trees \cite{cramer,koza} is frequently
called genetic programming.  The function trees are generated from a set of
17 primitives: the binary functions ADD, MULTIPLY and DIVIDE, and 14
terminals corresponding to the 14 indicators listed in
Table~\ref{marker-table}.  This set of primitives was established
empirically; conditional functions, subtraction, and random constants
failed to change performance significantly.  The polarities for several
indicators were reversed according to the polarities of the weights
established by log-linear regression.  Because the genetic algorithm is
stochastic, each run may produce a different function tree.  Runs of the
genetic algorithm have a population size of 500, and end after 50,000 new
individuals have been evaluated.

A threshold must be selected for both linear and function tree combinations
of indicators.  This way, overall outputs can be discriminated such that
classification performance is maximized.  For both methods, this threshold
is established over the training set and frozen for evaluation over the
test set.

\noindent
{\bf Decision trees.} Another method capable of modeling non-linear
relationships between indicators is a decision tree.  Each internal node of
a decision tree is a choice point, dividing an individual indicator into
ranges of possible values.  Each leaf node is labeled with a classification
(state or event).  Given the set of indicator values corresponding to a
verb, that verb's class is established by deterministically traversing the
tree from the root to a leaf.  The most popular method of decision tree
induction, employed here, is recursive partitioning \cite{quinlan,breiman},
which expands the tree from top to bottom.  The Splus statistical package
was used for the induction process, with parameters set to their default
values.

Previous efforts in corpus-based natural language processing have
incorporated machine learning methods to coordinate multiple linguistic
indicators, e.g., to classify adjectives according to markedness
\cite{Hatzivassiloglou&McKeown95}, to perform accent restoration
\cite{yarowsky2}, for disambiguation problems \cite{yarowsky2,luk95}, and
for the automatic identification of semantically related groups of words
\cite{Pereira&al93,Hatzivassiloglou&McKeown93}.  For more detail on the
machine learning experiments described here, see Siegel \shortcite{siegel7}.

\subsection{A Parsed Corpus}
\label{corpus}

The automatic identification of individual constituents within a clause is
necessary to compute the values of the linguistic indicators in
Table~\ref{marker-table}.  The English Slot Grammar (ESG) \cite{mccord} has
previously been used on corpora to accumulate aspectual data
\cite{klavans}.  ESG is particularly attractive for this task since its
output describes a clause's deep roles, detecting, for example, the deep
subject and object of a passivized phrase.

Our experiments are performed across a 1,159,891 word corpus of medical
discharge summaries from which 97,973 clauses were parsed fully by ESG,
with no self-diagnostic errors (ESG produced error messages on some of this
corpus' complex sentences).  The values of each indicator in
Table~\ref{marker-table} are computed, for each verb, across these 97,973
clauses.

\begin{table}
\begin{center}
\begin{tabular}{|l|r|r|r|}	\hline
%&&& \\
&n&States&Events \\
\hline
{\it be} & 23,409 & 100.0\% & 0.0\%  \\
{\it have} & 7,882 & 69.9\% & 30.1\%  \\
all other verbs & 66,682 & 16.2\% & 83.8\%  \\
% overall states  (.162 * 66682 + 31291) / 97973
% overall events  (.838 * 66682) / 97973
% overall accuracy  (23409+7882+(.838*66682))/97973
% dec tree overall accuracy ((.939*66682)+23409+7882)/97973
\hline
\end{tabular}
\end{center}
\caption{Breakdown of verb occurrences.}
\label{be-have-table}
\end{table}

In this paper, we evaluate our approach over verbs other than {\it be} and
{\it have}, the two most frequent verbs in this corpus.
Table~\ref{be-have-table} shows the distribution of clauses with {\it be},
{\it have}, and remaining verbs as their main verb.  Clauses with {\it
be} as their main verb always denote states.  
{\it Have} is highly ambiguous, so the aspectual classification of
clauses headed by {\it have} must incorporate additional constituents.  For
example, {\it ``The patient had Medicaid''} denotes a state, while, {\it
``The patient had an enema''} denotes an event.  In separate work, we have
shown that the semantic category of the direct object of {\it have} informs
classification according to stativity \cite{siegel7}.  Since the remaining
problem is to increase the classification accuracy over the 68.1\% of
clauses that have main verbs other than {\it be} and {\it have}, all results are
measured only across that portion of the corpus.  As shown in
Table~\ref{be-have-table}, 83.8\% of clauses with verbs other than {\it be}
and {\it have} are events.

%next-most-pop is ``show'', with 2131 occurrences

A portion of the parsed clauses must be manually classified to provide
supervised training data for the three learning methods mentioned above,
and to provide a separate set of test data with which to evaluate the
classification performance of our system.  To this end, we manually marked
1,851 clauses selected uniformly from the set of parsed clauses not headed
by {\it be} or {\it have}.  As a linguistic test to mark according to
stativity, each clause was tested for readability with {\it ``What happened
was...''}\footnote{This test was suggested by Judith Klavans (personal
communication).}  Of these, 373 were rejected because of parsing problems
(verb or direct object incorrectly identified).  This left 1,478 parsed
clauses, which were divided equally into 739 training and 739 testing
cases.

Some verbs can denote both states and events, depending on other
constituents of the clause.  For example, {\it show} denotes a state in
{\it ``His lumbar puncture showed evidence of white cells,''} but denotes
an event in {\it ``He showed me the photographs.''}  However, in this
corpus, most verbs other than {\it have} are highly dominated by one sense.
Of the 739 clauses included in the training set, 235 verbs occurred.  Only
11 of these verbs were observed as both states and events.  Among these,
there was a strong tendency towards one sense.  For example, {\it show}
appears primarily as a {\bf state}.  Only five verbs - {\it say}, {\it
state}, {\it supplement}, {\it describe}, and {\it lie}, were not dominated
by one class over 80\% of the time.  Further, each of these were observed
less than 6 times a piece, which makes the estimation of sense dominance
inaccurate.  

The limited presence of verbal ambiguity in the test set does, however,
place an upper bound of 97.4\% on classification accuracy, since linguistic
indicators are computed over the main verb only.

\section{Results}
\label{baba}

Since we are evaluating our approach over verbs other than {\it be} and
{\it have}, the test set is only 16.2\% states, as shown in
Table~\ref{be-have-table}.  Therefore, simply classifying every verb as an
event achieves an accuracy of 83.8\% over the 739 test cases, since 619 are
events.  However, this approach classifies all stative clauses incorrectly,
achieving a {\it stative recall} of 0.0\%.  This method serves as a
baseline for comparison since we are attempting to improve over an {\it
uninformed} approach.\footnote{Similar baselines for comparison have been
used for many classification problems \cite{duda}, e.g., part-of-speech
tagging \cite{church,allen}.}

\subsection{Individual Indicators}

The second and third columns of Table~\ref{marker-table} show the average
value for each indicator over stative and event clauses, as measured over
the 739 training examples.  As described above, these examples exclude {\it
be} and {\it have}.  For example, 4.44\% of stative clauses are modified by
either {\it not} or {\it never}, but only 1.56\% of event clauses were
modified by these adverbs.  The fourth column shows the results of T-tests
that compare the indicator values over stative verbs to those over event
verbs.  For example, there is less than a 0.05\% chance that the difference
between stative and event means for the first four indicators listed is due
to chance.  Overall, this shows that the differences in stative and event
averages are statistically significant for the first seven indicators
listed (p $<$ .01).

This analysis has revealed correlations between verb class and five
indicators that have not been linked to stativity in the linguistics
literature.  Of the top seven indicators shown to have positive
correlations with stativity, three have been linguistically motivated, as
shown in Table ~\ref{klavans-table1}.  The other four were not previously
hypothesized to correlate with aspectual class: (1) verb {\tt frequency},
(2) occurrences modified by {\it ``not''} or {``never''}, (3) occurrences
with no deep subject, and (4) occurrences in the past or present
participle.  Furthermore, the last of these seven, occurrences in the
perfect tense, was not previously hypothesized to correlate with stativity
in particular.

However, a positive correlation between indicator value and verb class does
not necessarily mean an indicator can be used to increase classification
accuracy.  Each indicator was tested individually for its ability to
improve classification accuracy over the baseline by selecting the best
classification threshold over the training data.  Only two indicators, verb
{\tt frequency}, and occurrences with {\it not} and {\it never}, were able
to improve classification accuracy over that obtained by classifying all
clauses as events.  To validate that this improved accuracy, the thresholds
established over the training set were used over the test set, with
resulting accuracies of 88.0\% and 84.0\%, respectively.  Binomial tests
showed the first of these to be a significant improvement over the baseline
of 83.8\%, but not the second.

\subsection{Combining Indicators}

All three machine learning methods successfully combined indicator values,
improving classification accuracy over the baseline measure.  As shown in
Table~\ref{results}, the decision tree's accuracy was 93.9\%, genetic
programming's function trees had an average accuracy of 91.2\% over seven
runs, and the log-linear regression achieved an 86.7\% accuracy.  Binomial
tests showed that both the decision tree and genetic programming achieved a
significant improvement over the 88.0\% accuracy achieved by the {\tt
frequency} indicator alone.  Therefore, we have shown that machine learning
methods can successfully combine multiple numerical indicators to improve
the accuracy by which verbs are classified.

The differences in accuracy between the three methods are each significant
(p $<$ .01).  Therefore, these results highlight the importance of how
linear and non-linear interactions between numerical linguistic indicators
are modeled.

%\begin{table}[b]
\begin{table*}
\begin{center}
\begin{tabular}{|l|r||rr||rr||} \hline
&overall&\multicolumn{2}{c||}{States}&
\multicolumn{2}{c||}{Events}\\
\cline{3-6}
&accuracy&recall&precision&recall&precision\\
\hline
\hline
decision tree & 93.9\% & 74.2\% & 86.4\% & 97.7\% & 95.1\% \\
\hline
genetic programming & 91.2\% & 47.4\% & 97.3\% & 99.7\% & 90.7\% \\
\hline
log-linear & 86.7\% & 34.2\% & 68.3\% & 96.9\% & 88.4\% \\
\hline
\hline
baseline & 83.8\% & 0.0\% & 100.0\% & 100.0\% & 83.8\% \\
\hline
\end{tabular}
\end{center}
\caption{Comparison of three learning methods and a performance baseline.}
\label{results}
\end{table*}

\subsection{Improved Recall Tradeoff}

The increase in the number of stative clauses correctly classified,
i.e. stative recall, illustrates a more dramatic improvement over the
baseline.  As shown in Table~\ref{results}, stative recalls of 74.2\%,
47.4\% and 34.2\% were achieved by the three learning methods, as compared
to the 0.0\% stative recall achieved by the baseline, while only a small
loss in recall over event clauses was suffered.  The baseline does not
classify any stative clauses correctly because it classifies all clauses as
events.  This difference in recall is more dramatic than the accuracy
improvement because of the dominance of event clauses in the test set.

This favorable tradeoff between recall values presents an advantage for
applications that weigh the identification of stative clauses more heavily
than that of event clauses.  For example, a prepositional phrase denoting a
duration with {\it for}, e.g., {\it ``for a minute,''} describes the
duration of a state, e.g., {\it ``She felt sick for two weeks,''} or the
duration of the state that results from a telic event, e.g., {\it ``She
left the room for a minute.''}  That is, correctly identifying the use of
{\it for} depends on identifying the stativity of the clause it modifies.
A language understanding system that incorrectly classifies {\it ``She felt
sick for two weeks''} as a non-telic event will not detect that {\it ``for
two weeks''} describes the duration of the {\it feel}-state.  If this
system, for example, summarizes durations, it is important to correctly
identify states.  In this case, our approach is advantageous.

\section{Conclusions and Future Work}

We have compiled a set of fourteen quantitative linguistic indicators that,
when used together, significantly improve the classification of verbs
according to stativity.  The values of these indicators are measured
automatically across a corpus of text.

Each of three machine learning techniques successfully combined the
indicators to improve classification performance.  The best of the three,
decision tree induction, achieved a classification accuracy of 93.9\%, as
compared to the uninformed baseline's accuracy of 83.8\%.  Furthermore,
genetic programming and log-linear regression also achieved improvements
over the baseline.  These results were measured over an unrestricted set of
verbs.

The improvement in classification performance is more dramatically
illustrated by the favorable tradeoff between stative and event recall
achieved by all three of these methods, which is profitable for tasks that
weigh the identification of states more heavily than events.

This analysis has revealed correlations between stativity and five
indicators that are not traditionally linked to stativity in the linguistic
literature.  Furthermore, one of these four, verb frequency, individually
increased classification accuracy from the baseline method to 88.0\%.

To classify a clause, the current system uses only the indicator values
corresponding to the clause's main verb.  This procedure could be expanded
to incorporate rules that classify a clause directly from clausal features
(e.g., Is the main verb {\it show}, is the clause in the progressive?), or
by calculating indicator values over other clausal constituents in addition
to the verb \cite{siegel4,siegel7}.

Classification performance may also improve by incorporating additional
linguistic indicators, such as co-occurrence with {\it rate} adverbs, e.g.,
{\it quickly}, or occurrences as a complement of {\it force} or {\it
persuade}, as suggested by Klavans and Chodorow \shortcite{klavans}.
%persuade}, as suggested by Klavans and Chodorow \cite{klavans}.

\section*{Acknowledgments}

Kathleen R. McKeown was extremely helpful regarding the formulation of our
work and Judith Klavans regarding linguistic techniques.  Alexander
D. Chaffee, Vasileios Hatzivassiloglou, Dragomir Radev and Dekai Wu
provided many helpful insights regarding the evaluation and presentation of
our results.

This research is supported in part by the Columbia University Center for
Advanced Technology in High Performance Computing and Communications in
Healthcare (funded by the New York State Science and Technology
Foundation), the Office of Naval Research under contract N00014-95-1-0745
and by the National Science Foundation under contract GER-90-24069.

Finally, we would like to thank Andy Singleton for the use of his GPQuick
software.

\bibliography{/u/boat/evs/prop/prop}
\bibliographystyle{fullname}

\end{document}